\documentclass[10pt,twocolumn,letterpaper]{article}

\usepackage[pagenumbers]{cvpr} 

\usepackage{graphicx}
\usepackage{amsmath}
\usepackage{amssymb}
\usepackage{booktabs}

\usepackage{multirow}
\usepackage{algorithmic}
\usepackage{algorithm}
\usepackage{color}
\usepackage{colortbl}

\usepackage{url}
\usepackage{xspace}
\usepackage{xcolor}
\usepackage{blindtext}
\usepackage{indentfirst}
\usepackage{ragged2e}
\usepackage{makecell}
\usepackage{diagbox}
\usepackage{bm}
\usepackage{booktabs} 
\usepackage[misc]{ifsym}

\usepackage{caption}
\usepackage{graphicx}
\usepackage{booktabs}
\usepackage{blindtext}
\usepackage{mathtools}

\makeatletter
\DeclareRobustCommand\onedot{\futurelet\@let@token\@onedot}
\def\@onedot{\ifx\@let@token.\else.\null\fi\xspace}
\def\eg{\emph{e.g}\onedot,~} 
\def\ie{\emph{i.e}\onedot,~}

\def\etal{\emph{et al}\onedot}
\makeatother

\DeclareMathOperator{\hist}{Histogram}
\DeclareMathOperator{\fps}{FPS}
\DeclareMathOperator{\cluster}{ClusterCenter}

\usepackage[pagebackref,breaklinks,colorlinks]{hyperref}

\usepackage[capitalize]{cleveref}
\crefname{section}{Sec.}{Secs.}
\Crefname{section}{Section}{Sections}
\Crefname{table}{Table}{Tables}
\crefname{table}{Tab.}{Tabs.}

\def\cvprPaperID{} 
\def\confName{CVPR}
\def\confYear{2022}

\begin{document}
\title{Label-Free Model Evaluation with Semi-Structured Dataset Representations}
\author{Xiaoxiao Sun, Yunzhong Hou, Hongdong Li, Liang Zheng\\
Australian National University\\
{\tt\small \{first name.last name\}@anu.edu.au}
}
\maketitle

\begin{abstract}
Label-free model evaluation, or AutoEval, estimates model accuracy on unlabeled test sets, and is critical for understanding model behaviors in various unseen environments. In the absence of image labels, based on dataset representations, we estimate model performance for AutoEval with regression. On the one hand, image feature is a straightforward choice for such representations, but it hampers regression learning due to being unstructured (\ie no specific meanings for component at certain location) and of large-scale. On the other hand, previous methods adopt simple structured representations (like average confidence or average feature), but insufficient to capture the data characteristics given their limited dimensions. In this work, we take the best of both worlds and propose a new semi-structured dataset representation that is manageable for regression learning while containing rich information for AutoEval. Based on image features, we integrate distribution shapes, clusters, and representative samples for a semi-structured dataset representation. Besides the structured overall description with distribution shapes, the unstructured description with clusters and representative samples include additional fine-grained information facilitating the AutoEval task. On three existing datasets and 25 newly introduced ones, we experimentally show that the proposed representation achieves competitive results. Code and dataset are available at \url{https://github.com/sxzrt/Semi-Structured-Dataset-Representations}.
\end{abstract}

\section{Introduction}
\label{sec:intro}
Model evaluation is critical for machine learning and computer vision, which usually requires a test set with ground truth labels. For real-world applications, however, once the model is deployed, conducting standard evaluation can be tricky: data collected on the fly are unlabeled, but additional human labeling for each different application scenario can be extremely costly and time-consuming. To evaluate model performance on various unmet data distributions, researchers investigate label-free model evaluation, or AutoEval, which predicts model accuracy based on only unlabeled test sets~\cite{deng2021labels, garfield2002challenge, deng2021does, chen2021detecting}.

\begin{figure}[t]
\begin{center}
    \includegraphics[width=\linewidth]{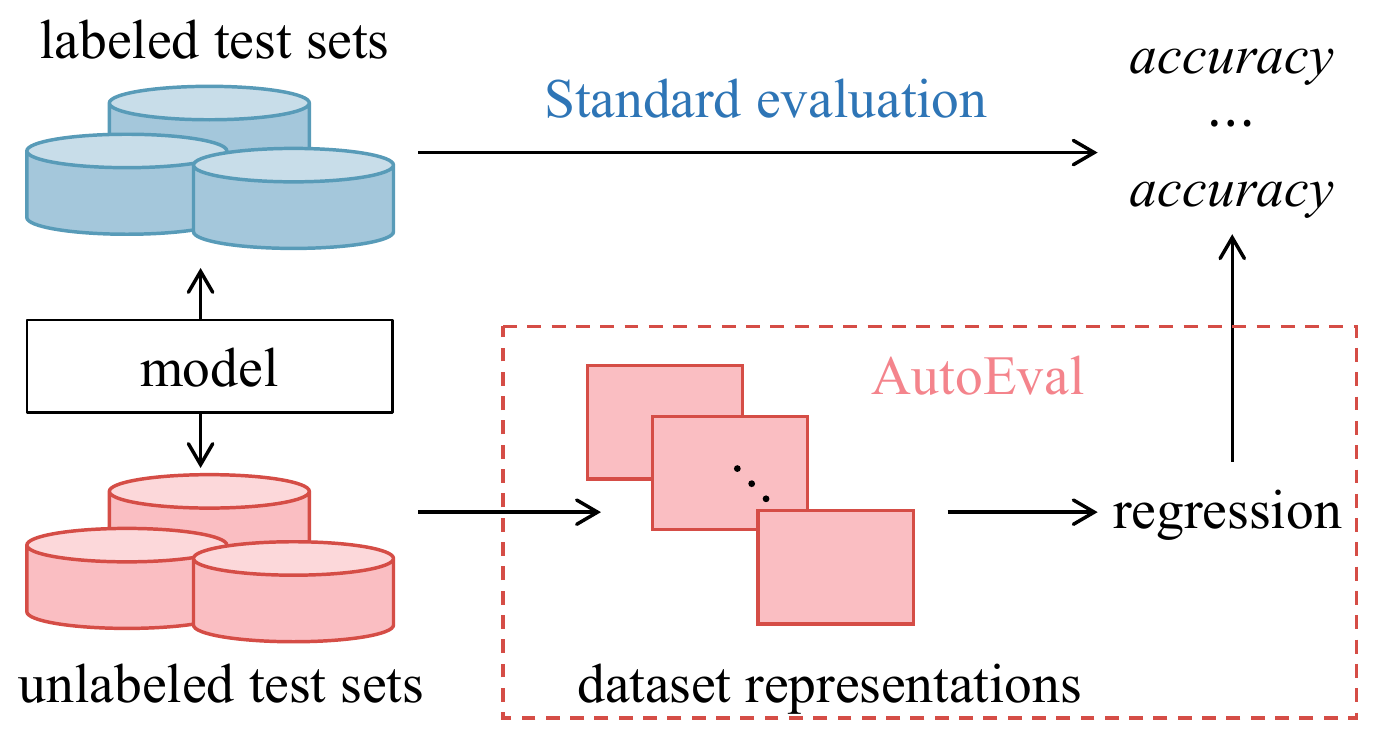}
\end{center}
\caption{
Illustration of standard model evaluation and label-free model evaluation (AutoEval). 
\textbf{Top}: Standard evaluation calculates model accuracy based on model outputs and corresponding test labels. 
\textbf{Bottom}: AutoEval, on the other hand, has no access to test labels. However, based on dataset representations, we can estimate the model accuracy using regression methods. 
}
\label{fig:fig1}
\end{figure}

In order to estimate model accuracy without ground truth labels, one can rely on certain representations that can encode both the model and the dataset, and then regress the accuracy based on that representation. We later on denote such representations as dataset representations. As shown in Fig.~\ref{fig:fig1}, based on dataset representation, we can learn to estimate the model performance as a regression problem.

\begin{figure*}[t]
\begin{center}
\includegraphics[width=\linewidth]{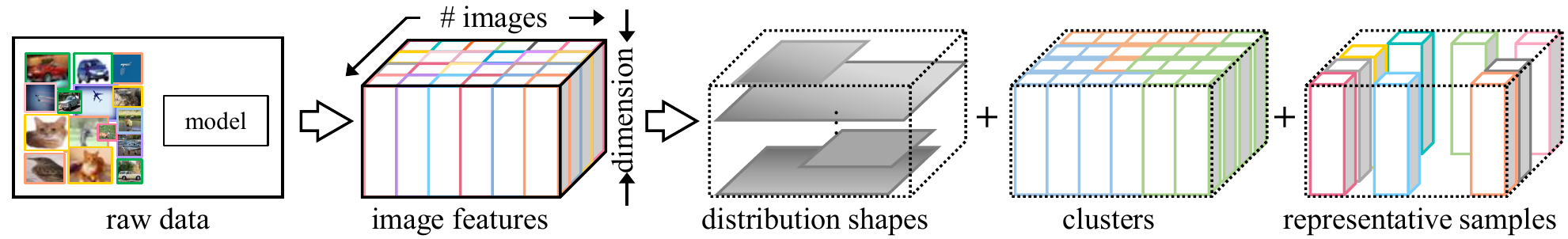}
\end{center}
\caption{The proposed semi-structured dataset representation and its three components. 
\textbf{Distribution shapes} (structured) encode the marginal distribution for each of the feature dimensions (Section~\ref{secsec:shape}). 
\textbf{Clusters} (unstructured) provide more fine-grained representation of the overall distribution (Section~\ref{secsec:clusters}).
\textbf{Representative samples} (unstructured) are selected to reflect the dataset variability (Section~\ref{secsec:samples}). 
}
\label{fig:fig-pip}
\end{figure*}

As for dataset representations, on one hand, image features are the most straightforward choice \cite{deng2021labels, garfield2002challenge}. 
However, there are problems with image features as dataset representations: extracted from each image, the combined set of features can be large-scale and unstructured, both of which bring troubles to regression learning in AutoEval. 
On the other hand, existing literature adopts simple structured dataset representation for AutoEval, which includes: 
1) statistics of prediction scores, such as the average confidence~\cite{guillory2021predicting} and entropy~\cite{saito2019semi} output by the last layer of the model; 
2) statistics of penultimate-layer features, such as the first and second-order statistics (mean and variance) of image features~\cite{peng2019moment, scholkopf2007kernel}, distribution distance between the training and target test sets (\eg Fr\'{e}chet distance (FD)~\cite{deng2021labels,heusel2017gans}, Maximum Mean Discrepancy~\cite{borgwardt2006integrating});
and 3) model performance on related pretext task~\cite{deng2021does, gidaris2018unsupervised}, such as rotation prediction accuracy. 
Nonetheless, such representations also have their problems as their limited dimensions (\eg feature distance or model performance on related tasks can be represented as a number, and feature mean only takes up a vector) might contain too little information to characterize the complex dataset distribution. 
Facing these problems, we investigate how to get a dataset representation that contains abundant information while remaining manageable for regression learning in AutoEval.

In this paper, we propose a new semi-structured representation that contains distribution shapes, clusters, and representative samples. We show a demonstration of the proposed representation in Fig.~\ref{fig:fig-pip}. 
\textit{First}, we include a structured descriptor with distribution shapes. Since the joint distribution across all feature channels can be very difficult to specify, we consider the marginal distributions of the image features. More specifically, for each dimension, we consider their marginal distribution, and record their probability density functions.
\textit{Second}, we include an unstructured descriptor with clusters. Since the global statistics can be too coarse and may fail to specify some local structures, we consider representing an overall distribution with its clusters. We use centers to represent the clusters of image features, which can reflect more fine-grained information of distribution, such as cluster locations and inter-cluster distances. 
\textit{Third}, we include another unstructured descriptor with representative samples. Specifically, we adopt Farthest Point Strategy ($\fps$) to sample representative image features to reflect dataset variability. 
Combining the three components, we have an overall semi-structured representation that takes the best of both worlds: manageable for regression learning (unlike raw image features), while at the same time containing richer information than previous simple structured representations.

To verify the effectiveness of the proposed method, we learn regression models and evaluate on AutoEval tasks. For both digit recognition and image classification, on three existing real test sets, their synthetic variants, and 25 newly introduced real test sets, we witness consistent performance improvements over existing baselines.

The main contributions of this paper include:
\begin{itemize}
  \item We propose a new semi-structured dataset representation from image features. Integrating distribution shape, clusters, and representative samples, the proposed representation encodes abundant information for AutoEval tasks. 
  \item We collect 25 real-world evaluation datasets from websites for more comprehensive evaluation in AutoEval. Experimental results on 3 existing sets, their synthetic variants, and 25 newly introduced web sets show that our semi-structured dataset representation can lead to better estimation of system performance.
\end{itemize}

\section{Related Work}
\label{sec:related}

\textbf{Date-centric AutoEval} predicts performance of a certain model on \textit{different test sets}~\cite{deng2021does, deng2021labels, guillory2021predicting, zilly2019frechet}. In a related field, model-centric AutoEval, or model generalization, predicts the performance of \textit{different models} on the same environment \cite{jiang2018predicting,neyshabur2017exploring}.
In this paper, we focus primarily on data-centric AutoEval, and shorten it as AutoEval if not specified. 
For this task, researchers generally use ``output'' (image features or prediction scores on a dataset) of the classification model to train a regression model that models the relationship between test set/training set similarity and the accuracy. 
For example, Deng \etal~\cite{deng2021labels} use Fr\'{e}chet distance and distribution-related feature statistics (mean and covariance) to characterize a dataset set. Guillory \etal~\cite{guillory2021predicting} think that the difference of confidences (DoC) is informative over various forms of distribution shift and use it as a dataset representation for AutoEval. Apart from using features to get dataset representation, Deng \etal~\cite{deng2021does} use rotation prediction on the test set to regress the classification accuracy on it. We speculate a single scalar or vector possesses too limited discriminative ability to describe a complex dataset, and this limitation motivates us to design more effective dataset representations. 

\textbf{Model-centric AutoEval}, also known as model generalization~\cite{corneanu2020computing,neyshabur2017exploring, dziugaite2020search}, predicts the generalization gap (the gap between the training and testing accuracy) of \textit{various models}. Some works~\cite{jiang2018predicting, jiang2019fantastic} aim to close this gap by developing bounds on generalization error, optimization error, and excess risk. For example, Jiang \etal \cite{jiang2019fantastic} conduct a large-scale empirical study aimed at uncovering potential causal relationships between bounds and generalization. Note that, these methods are proposed based on an assumption that the training and test distributions are the same. So, it is very different with our study because we focus on predicting model performance on test sets from various distributions. 

\textbf{Out-of-distribution detection} (OOD) detects test samples from a different distribution than the training distribution \cite{devries2018learning, lee2017training, Zhang_2021_CVPR, Lin_2021_CVPR}. OOD is related to data-centric AutoEval tasks in the sense that former evaluates data samples individually while the latter considers the distribution on the whole. 
For this individual evaluation task, some works use the outputs of the model, such as maximum softmax probability ~\cite{hendrycks2016baseline}, energy score~\cite{liu2020energy}, and confidence score estimated from a special branch in the model~\cite{devries2018learning}, as clues to judge out-of-distribution data. However, they do not consider the overall distribution of a set. Differently, we use statistical strategies to qualify the representation of a distribution rather than special samples from a distribution, which is more challenging than describing a single image because a set contains much more information.

\textbf{Set representation} has been investigated in some machine learning tasks~\cite{zaheer2017deep, skianis2020rep,kusner2015word, qi2017pointnet, naderializadeh2021set}, ranging from population statistic estimation~\cite{poczos2013distribution}, point cloud classification~\cite{qi2017pointnet, li2018so} and outlier detection~\cite{zaheer2017deep}. They often focus on designing neural network architectures that can handle sets. For example, Naderializadeh \etal~\cite{naderializadeh2021set} propose a framework for learning representations from set-structured data based on Generalized Sliced-Wasserstein Embedding.  Lee \etal~\cite{lee2019set} devise a neural network called Set Transformer, that uses self-attention to model interactions among the elements of the input set. The research of these works focuses on the representation of the set itself, while our study involves designing a representation related to both the dataset and the classification model to be evaluated.

\section{Problem Formulation}
Given a set of images $\mathcal{X}=\{\bm{x}\}$ and its corresponding label set $\mathcal{Y}=\{y\}$, standard evaluation can output the accuracy of a certain model $f$ as $Acc(f,\mathcal{X},\mathcal{Y})$.
However, for real-world deployments, oftentimes there are no available human-annotated labels. In this case, label-free model evaluation, or AutoEval, estimates the model performance in the absence of label set $\mathcal{Y}$, 
\begin{equation}
\label{eq:autoeval}
\widetilde{Acc}(f,\mathcal{X}) \rightarrow Acc(f,\mathcal{X},\mathcal{Y}),
\end{equation}
where $\rightarrow$ denotes the approximation goal. 

In order to provide such an approximation, one have to rely on the information available: model $f$ and image set $\mathcal{X}$. A most straightforward way of representing both the model and the unlabeled dataset is the image features for that specific model and dataset. 
Using the feature $f(\bm{x})$ (extracted from a certain layer in model $f$) of image $\bm{x}\in \mathcal{X}$, we can encode information of both the model $f$ and the dataset $\mathcal{X}$ into a set of image features 
\begin{equation}
\label{eq:image_feat}
\mathcal{F} = \{f(\bm{x})\}, \qquad \bm{x}\in \mathcal{X}.
\end{equation}

However, the image feature set still has several problems to facilitate the AutoEval task in Eq.~\ref{eq:autoeval}. More concretely, the set of image features  $\mathcal{F}$ is still large-scale, orderless, and highly complex. In this case, we aim to provide a semi-structured dataset representation of the feature set $\mathcal{F}$,
\begin{equation}
\label{eq:strcuctured_rep}
\mathcal{F}\xRightarrow[]{\text{representation}}\bm{h},
\end{equation}
which is lower-dimensional and has specific meanings for a part of locations. Such a dataset representation $\bm{h}$ should also encode abundant information for AutoEval, since it is our goal to learn a regression model $g(\cdot)$ that can estimate the accuracy of model $f$ on dataset $\mathcal{X}$ for AutoEval,
\begin{equation}
\label{eq:regress}
\widetilde{Acc}(f,\mathcal{X}) = g(\bm{h}).
\end{equation}
In our experiments, we adopt the fully connected design for the regression network $g$ following Deng \etal \cite{deng2021does}. 



Lastly, we can train the regression model $g$ on labelled sample sets $\left<\mathcal{X}_1, \mathcal{Y}_1\right>, \left<\mathcal{X}_2, \mathcal{Y}_2\right>, \cdots$ that are generated by transformations. For a certain model $f$, we can extract different sets of image features $\mathcal{F}_1, \mathcal{F}_2, \cdots$ from the image sets $\mathcal{X}_1, \mathcal{X}_2, \cdots$ using Eq.~\ref{eq:image_feat}, and then convert them into semi-structured representation of the feature sets $\bm{h}_1, \bm{h}_2,  \cdots$ using Eq.~\ref{eq:strcuctured_rep}. Combined with the ground truth accuracies, we finally create \textit{support sets} $\left<\bm{h}_1, Acc(f,\mathcal{X}_1,\mathcal{Y}_1)\right>, \left<\bm{h}_2, Acc(f,\mathcal{X}_2,\mathcal{Y}_2)\right>,  \cdots$ for the regression learning of $g$ using Eq.~\ref{eq:autoeval} and Eq.~\ref{eq:regress}.

\section{Semi-Structured Dataset Representation}
\label{sec:method}
This paper studies dataset representation for AutoEval task. 
Based on the set of image features $\mathcal{F}$, we focus on how to design a semi-structured dataset representation $\bm{h}$. Specifically, the proposed dataset representation consists of three components: distribution shape, clusters, and representative samples. The overall representation is shown in Fig.~\ref{fig:fig-pip}. In the following sections, we give a detailed description of each of the components.

\begin{figure}[t]
\begin{center}
	\includegraphics[width=\linewidth]{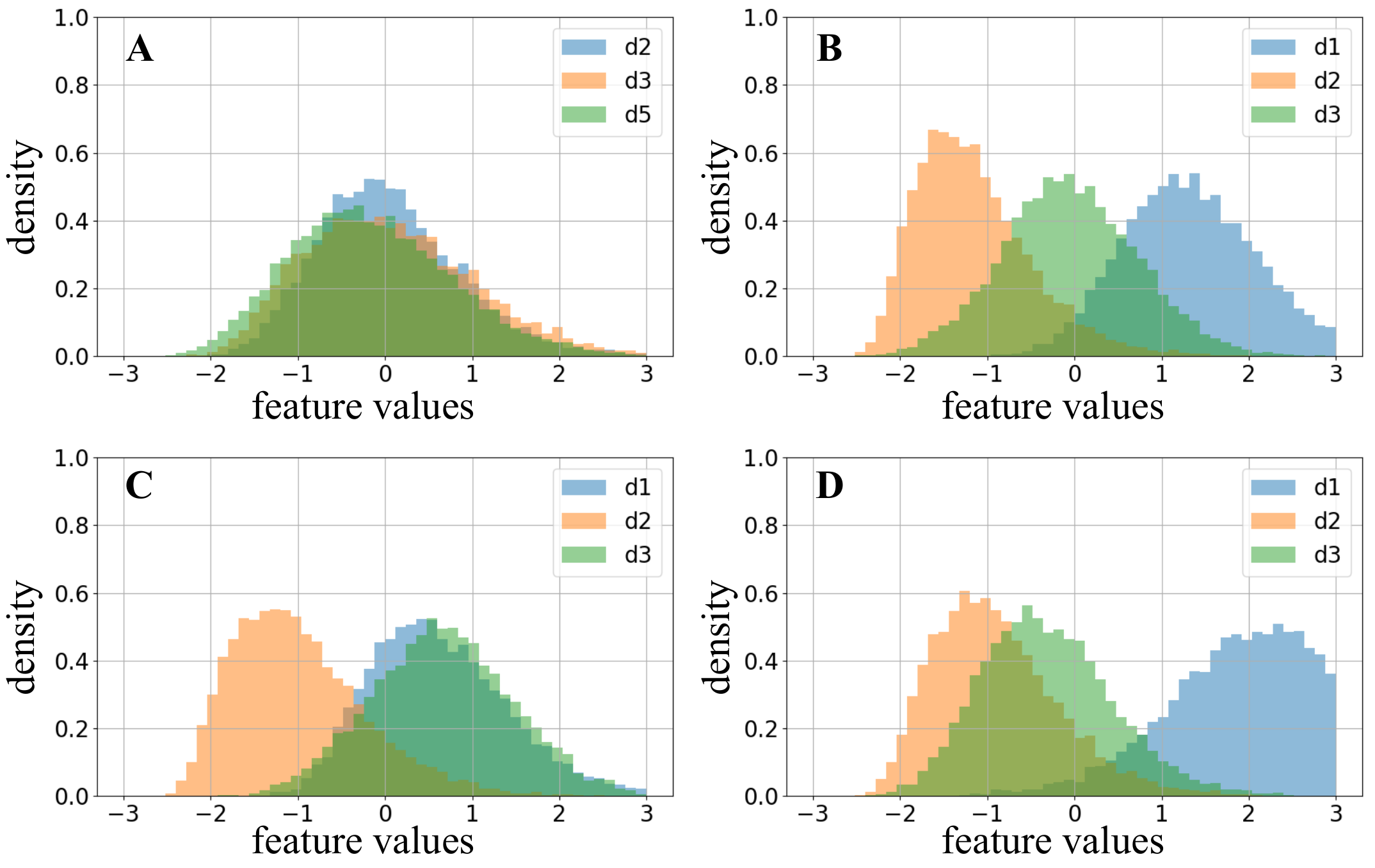}
\end{center}
\caption{Examples of shape representation $\bm{h}_\text{shape}$ for four datasets (\textbf{A}, \textbf{B}, \textbf{C}, and \textbf{D}). Three-dimensional (d1, d2 and d3) image features are used. 
In each subfigure, we show the proposed distribution shape representation $\bm{h}_\text{shape}$ as the combination of marginal distribution histograms. 
We can observe that 1) for each dataset, its feature histograms for the three dimensions are different, and that 2) the four datasets have distinctive histograms of features.}
\label{fig:fig2}
\end{figure}


\subsection{Distribution Shape} 
\label{secsec:shape}

A one-dimensional distribution can be characterized by its probability density function. In the digital world of computers, we can use histograms as a discretized version of the probability density function. Such representations are also known as the shape of the distributions. 

For multi-dimensional distributions like the set of image features $\mathcal{F}$, recording the probability density function or the distribution shape across multiple dimensions can be difficult. In this case, we consider only the marginal distributions, \ie the distribution for each of the dimensions in the image feature. Moreover, different dimensions of neural network features are believed to record different semantic information~\cite{bau2017network}. 
For $D$-dimensional image features in $\mathcal{F}$, we record histograms of $D$ marginal distributions, and then combine them together to formulate a \textit{structured} descriptor,
\begin{equation}
\bm{h}_\text{shape} = \hist(\mathcal{F},B) \in \mathbb{R}^{D\times B},
\end{equation}
where $\hist(\cdot, \cdot)$ denotes the combination of histograms from $D$ marginal distributions, $B$ denotes the number of bins in histograms. The resulting distribution shape representation $\bm{h}_\text{shape}$ is therefore $D\times B$ dimensional. 
For easier understanding, we show some examples of the proposed distribution shape representation $\bm{h}_\text{shape}$ in Fig.~\ref{fig:fig2}.

Although effective, representing the distribution shape with marginal distributions also has its limitations. The most prominent one is that the resulting representation does not encode any correlation information for different dimensions, \ie the covariance matrix from the proposed shape representation will only have non-zero terms on its diagonal. Therefore, we additionally include distribution clusters and representative samples, two \textit{unstructured} descriptors in our semi-structured dataset representation.  

\begin{figure}[t]
\begin{center}
	\includegraphics[width=\linewidth]{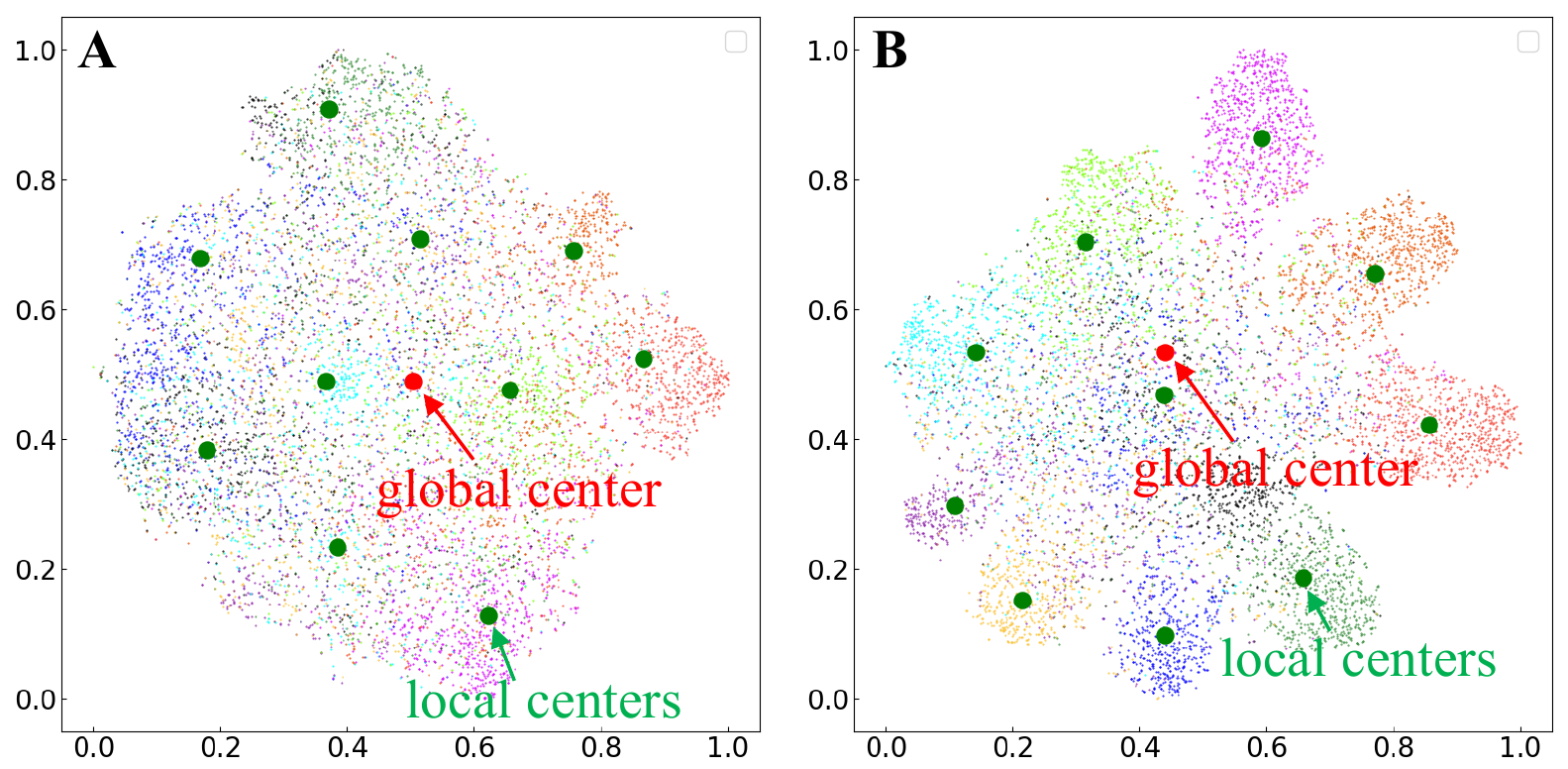}
\end{center}
\caption{Illustration of global center and local centers. \textbf{A} and \textbf{B} visualize feature distributions two synthetic datasets under the CIFAR-10 setup. Red and green points represent the global center and cluster centers, respectively. T-SNE~\cite{van2008visualizing} is used.}
\label{fig:fig3}
\end{figure}

\subsection{Cluster}
\label{secsec:clusters}

Rather than representing a distribution in an overall manner \cite{deng2021labels, peng2019moment, scholkopf2007kernel}, in this work, we additionally represent the distribution with its parts to include more fine-grained information. For example, in Fig.~\ref{fig:fig3}, global centers of two sets are similar, but their distributions are very different. 

Since we primarily focus on classification problems in this work, we represent the parts of the overall distribution as $K$ clusters, where $K$ is the number of classes considered in the model. As shown in Fig.~\ref{fig:fig3}, for $K=10$ class recognition problems, its image features can roughly be clustered into 10 clusters. The distribution of these 10 clusters can provide additional information on how the model $f$ adapts to dataset $\mathcal{X}$, further benefiting the AutoEval task. Hence, we propose cluster representation as follows. Given image features in $\mathcal{F}$, we group them into $K$ clusters, and use the cluster centers as \textit{unstructured} fine-grained representation,
\begin{equation} \small
\bm{h}_\text{cluster} = \cluster(\mathcal{F},K) \in \mathbb{R}^{D\times K},
\end{equation}
where $\cluster(\cdot,\cdot)$ denotes the cluster centers. The resulting cluster representation is $D\times K$ dimensional.

\subsection{Representative Samples} 
\label{secsec:samples}
\begin{figure}[t]
\begin{center}
		\includegraphics[width=\linewidth]{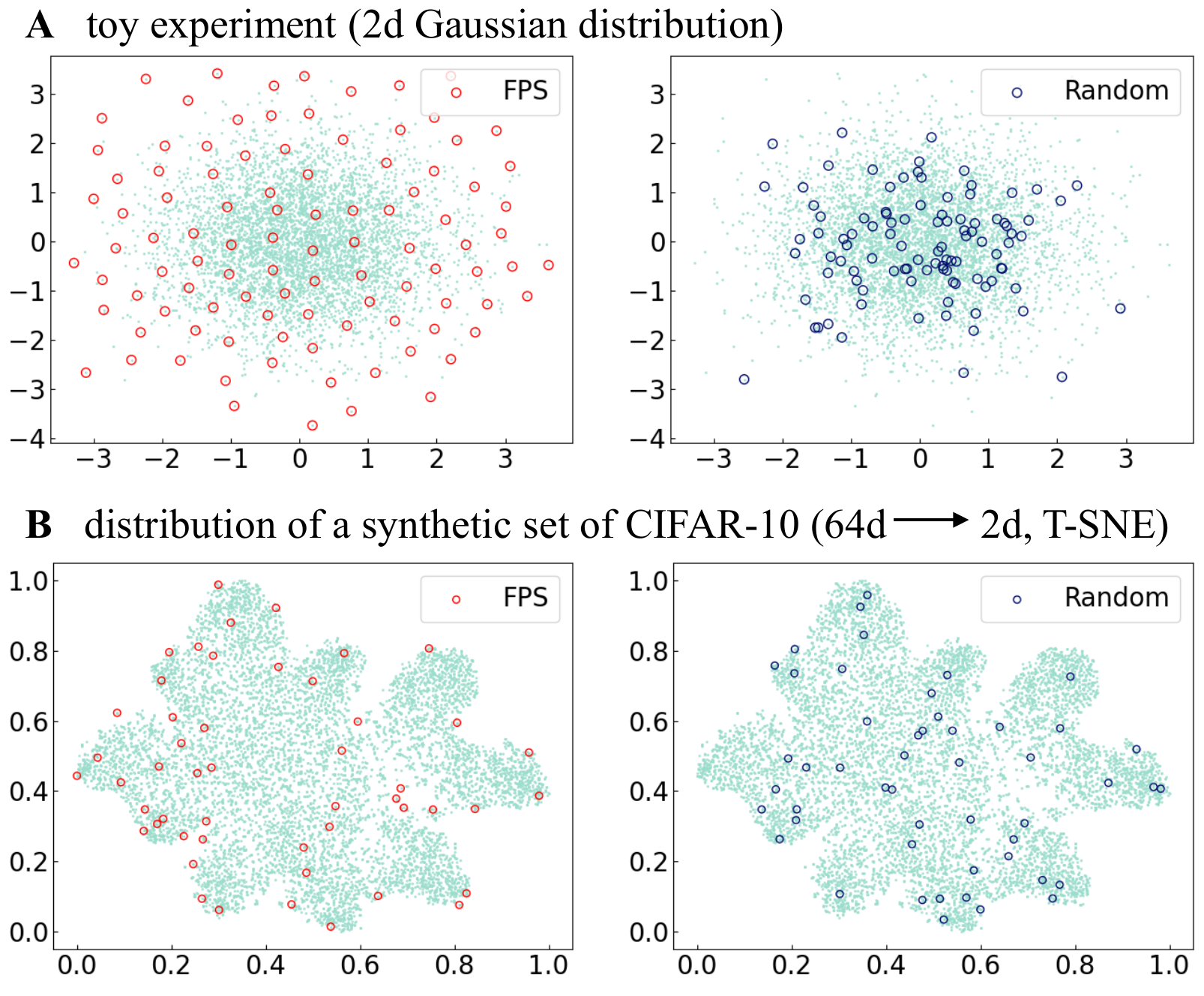}
\end{center}
\caption{Farthest points sampling ($\fps$) \textit{vs.} random sampling on \textbf{A} toy 2-dim Gaussian distribution and \textbf{B} a real dataset distribution. $\fps$ allows us to reach the distribution borders, and thus depict the distribution spread, whereas random sampling is entirely based on the distribution (probability density function). 
}
\label{fig:fig4}
\end{figure}

In addition to distribution shapes and centers, we also represent data variability with representative samples. Feature variance, another form for variability representation \cite{jonsson1982some, deng2021labels}, is already encoded in the proposed distribution shape vector $\bm{h}_\text{shape}$.
In this case, we consider sampling ``data on the edge" from it with farthest points sampling ($\fps$)~\cite{eldar1997farthest}. $\fps$ starts with a random point, and iteratively selects the farthest point from the already selected ones. As shown in Fig.~\ref{fig:fig4}, unlike random sampling according to the distributions, $\fps$ covers most of the feature spanning and is regardless of distribution. Mathematically, we create a $D\times S$-dimensional \textit{unstructured} representation as,
\begin{equation}
\label{eq:fps}
\bm{h}_\text{sample} = \fps(\mathcal{F},S) \in \mathbb{R}^{D\times S},
\end{equation}
where $S$ denotes the number of samples considered. 
Since the feature distance calculation is based on all $D$ dimensions, $\fps$ results can to a certain extent encodes correlations between different dimensions that are not represented in our marginal distribution shapes (see Section~\ref{secsec:shape}). 

Combining all three components, we have the overall semi-structured dataset representation,
\begin{equation}
\label{eq:overall}
\bm{h} = [\bm{h}_\text{shape},\bm{h}_\text{cluster},\bm{h}_\text{sample}] \in \mathbb{R}^{D\times (B+K+S)},
\end{equation}
which can be used for regression model learning in Eq.~\ref{eq:regress}.

\subsection{Discussion}
\textbf{Application scope of dataset representations.} Except AutoEval, potentially there could be other applications that require dataset representations. Examples include estimating dataset noise level and predicting training set quality. The latter has been studied in some works~\cite{zendel2017analyzing, yan2020neural}, but only hand-crafted features such as FD and MMD are used. Moreover, when the task is different, dataset representations should be different, too. Even for AutoEval, dataset representations for the classification task should be different from those for dense prediction. In this regard, it would be interesting to further explore other dataset-level tasks and their dataset representations.

\textbf{AutoEval: model-centric versus data-centric.} We notice that there is a body of works focusing on predicting the performance of various models given fixed training and test sets~\cite{steinhardt2016unsupervised, platanios2017estimating}, or model-centric AutoEval. Our focus is different in that we are interested in the performance of a given model trained by a fixed training set, under various test sets, or data-centric AutoEval. Therefore, the former studies \textit{data-agnostic} model statistics, while we study \textit{dataset} representations based on image features from certain models. It would be interesting to combine both, as done in \cite{guillory2021predicting}. Nevertheless, since data-centric AutoEval is still at an early stage, our efforts are mainly made in this domain, and we will leave the combination of model-centric and data-centric AutoEval to future work.

\begin{figure}[t]
\begin{center}
	\includegraphics[width=\linewidth]{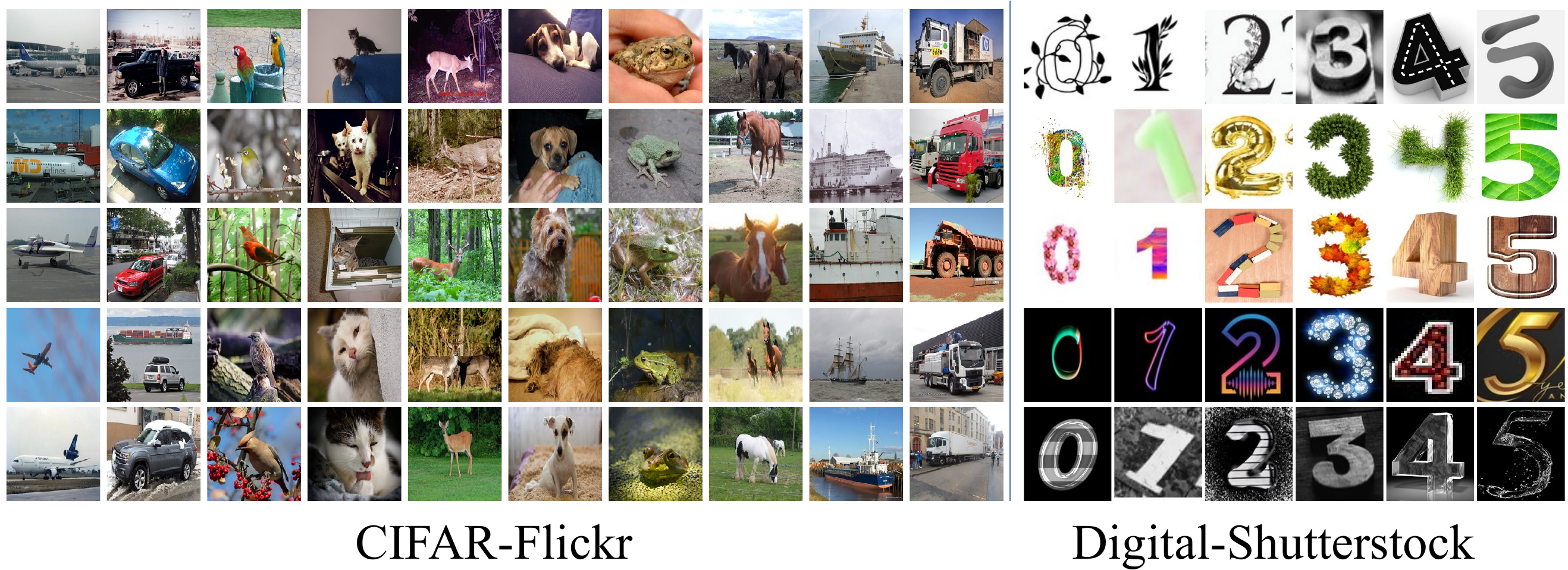}
\end{center}
\caption{Sample images of newly collected test sets, CIFAR-Flickr (left) and Digital-Shutterstock (right). The five rows of CIFAR-Flickr contain images sampled from five different years: 2011, 2012, 2013, 2018 and 2019. 
On the right, each row shows images of one of the five Digital-Shutterstock datasets. These newly collected real-world datasets are clearly very diverse.}
\label{fig:fig-example}
\end{figure}

\begin{table*}[t]
\begin{center}
\small
\setlength{\tabcolsep}{0.6mm}{
\begin{tabular}{l | c c c | c c c|c } 
\Xhline{1.2pt}
\multirow{2}{*}{Method}  &\multicolumn{3}{c|}{CIFAR-10} & \multicolumn{3}{c|}{MNIST} & \multicolumn{1}{c}{TinyImageNet}\\ 
 \cline{2-8}
 & CIFAR-10.1 (1) & CIFAR-C (30) & CIFAR--F (20) &  SVHN (1)  &  SVHN-C (30) &  Digital-S (5)   & TinyImageNet-C (75) \\ 
  \hline
Prediction score ($\tau_1$ = 0.8) & 3.4700  & 1.8217  & 8.2305  &	0.9400 &	2.1492	&10.5147 &  6.2130\\ 
Prediction score ($\tau_1$ = 0.9)  & 0.9000  & \textbf{1.2634}  & 7.5367 	& 3.3500 & 	2.5362	& 11.2857  & 8.5362 \\ 
  \hline
Entropy score ($\tau_2$ = 0.2) & 1.5500 & 1.7963  & 7.6894 &	3.3300 &	2.4795	&11.1798  & 8.4795 \\ 
Entropy score ($\tau_2$ = 0.3) & 6.1600 & 2.0131  & 8.8214  &	2.2600 &	2.3576 &	10.7841 &  6.3576\\ 
   \hline
FD \cite{deng2021labels} & 0.9600 & 1.9358 & 7.6635	 &0.8500 &	3.2964 &	11.9135 & 8.1325\\
AC or DOC \cite{guillory2021predicting} & 0.8400 & 1.5372 &  7.4107 &	0.8200 &	1.8963&	11.0411& 6.2647  \\
Rotation \cite{deng2021does}&  4.2400 & 1.9875  &  8.4881	& 2.4900 &2.7165	& 11.4772 & 8.2148\\
\hline
FD+$\sigma$+$\tau$ \cite{deng2021labels} & 0.8300 &  1.8324  & 7.6301 & 	2.0900 &	3.0418	& 11.7895 &  7.9635
 \\
\hline
Ours & \textbf{0.7400} & 1.2840 & \textbf{7.0213}  &	\textbf{0.7600} &	\textbf{1.8713} & 	\textbf{9.5204} & \textbf{5.9471}\\
\Xhline{1.2pt} 
    \end{tabular}}
\end{center}
\vspace{-8pt}
\caption{Comparison of various dataset representations in the AutoEval task. RMSE (\%) is used. ``Prediction score'', ``Entropy score'', ``FD'', ``AC'' (``or ODC'') and ``Rotation'' use a scalar to represent a dataset. Deng and Zheng~\cite{deng2021labels} uses FD, mean and variance calculated on images features as set representation. Numbers in brackets indicate the number of test sets. On CIFAR-F, Digital-S and TinyImageNet-C which contain many test sets, our system yields the lowest RMSE. } 
\label{table:Result}
\end{table*}

\section{Experiments}
\label{sec: experiments}

\subsection{Datasets}\label{sec:dataset}
\textbf{CIFAR-10 setup.}
We train the classifier model $f$ on the training set of CIFAR-10 \cite{krizhevsky2009learning}, which contains 50,000 32$\times$32 images across 10 classes. 
For AutoEval, we use 800 sample sets for regression learning of $g$ with Eq.~\ref{eq:regress}, which are generated from 
various geometric and photometric transformations on top of CIFAR-10 test set following previous work~\cite{deng2021labels}. 
(An additional 200 datasets is generated in similar manners for ablation and variant study. )
To evaluate the regression performance of AutoEval, we use 
three test set setups: (1) CIFAR-10.1, (2) CIFAR-C: 30 synthetics sets generated on CIFAR-10.1 with unseen transformations used in sample sets, and (3) CIFAR-Flickr (CIFAR-F): 20 real-word test sets \textit{newly} collected from Flickr by searching 10 keywords corresponding to 10 classes of the CIFAR-10 from 20 years (2001-2020). respectively. Fig.~\ref{fig:fig-example} shows some sample images of the newly collected sets. 

\textbf{MNIST setup.}
Similarly, we train the classifier model $f$ on training set of MNIST~\cite{deng2012mnist}, which contains 50,000 training image for 10 digits. 
Based on MNIST test set, we generate 800 sample sets on MNIST for regression learning of $g$, and 200 datasets for ablation study. 
However, during generating sample sets, a new operation is used on each image that is background replacement. Background images are sampled from COCO~\cite{lin2014microsoft} training sets following~\cite{deng2021labels}. The unseen test sets of this setup are: (1) SVHN, (2) SVHN-C: 30 synthetics sets generated on SVHN with unseen transformations used in sample sets, and (3) Digital-Shutterstock (Digital-S): 5 \textit{new} test sets that are searched from Shutterstock based on different options of color. Numbers (Fig.~\ref{fig:fig-example} right) are cropped from the original web image manually as data of Digital-Shutterstock.

\textbf{TinyImageNet setup.}
We train the classifier model $f$ on training partition of the TinyImageNet dataset, which contains 100,000 64$\times$64 colored images of 200 classes. 
Based on the publically available validation partition of TinyImageNet, we create 700 sample sets for regression learning of $g$ and 50 dataset variants for variant and ablation study. 
To evaluate regression performance for AutoEval, we use TinyImageNet-C:  75 sets generated on validation sets based on synthetic corruptions and perturbations present in ImageNet-C~\cite{hendrycks2019benchmarking}. During generating test sets, we take special care to ensure there is no overlapping between methods synthesizing sample and test sets.

\subsection{Experimental Settings}


\textbf{Classification models.}
Our classifier design follows \cite{deng2021labels}. Specifically, we train a 10-way classifier on the CIFAR-10 training set using. ResNet-44~\cite{he2016deep} as the backbone. For MNIST setup, we train a 10-way classifier on its training set, and the backbone is LeNet-5~\cite{lecun1998gradient}. Both classifiers are trained from scratch. A 200-way classifier is trained on the TinyImageNet training set. We use ResNet-50 on ImageNet~\cite{deng2009imagenet} as the pretrained model and fine-tune it on TinyImageNet. We additionally train a VGG-16 classifier for the CIFAR-10 setup.

\textbf{Metric.} To evaluate the effectiveness of classifier accuracy estimation, we use the root mean squared error (RMSE), which measures the average squared distance between the estimated accuracy and ground-truth accuracy. A small RMSE represents good estimation and vice versa. 

All experiments are conducted on one RTX-2080TI GPU and a 16-core AMD Threadripper CPU @ 3.5Ghz. 
 
\begin{table*} 
		\begin{minipage}{0.38\linewidth}\footnotesize
		\centering
		\vspace{3mm}
		    \setlength{\tabcolsep}{2mm}{
		\begin{tabular}{l | c| c| c}
        			\toprule
       \multirow{2}{*}{Method}  		  &  CIFAR-10.1 &  CIFAR-C &   CIFAR-F \\
       &  (1)  &  (30) &  (20)\\
        			\midrule
        FD~\cite{deng2021labels} & 0.9200 & 1.8553 & 7.5439\\
        AC~\cite{guillory2021predicting}& 0.7900 & 	1.4917 &	7.1420\\
        Rotation~\cite{deng2021does}  & 3.4800 & 2.1496 & 8.7212 \\
        FD+$\sigma$+$\tau$  & 0.8200& 1.5369& 7.2715\\
        \hline
Ours	& 0.6900 & 1.1830 & 6.5472 \\
			\bottomrule
		\end{tabular}}
	    \caption{AutoEval of the VGG-16 classifier using various dataset representations under the CIFAR-10 setup. The settings are same as those in Table~\ref{table:Result}, except that images features are extracted from VGG-16. RMSE (\%) is shown: lower is better.}
		\label{table:vgg-16}
	\end{minipage}
	\hfill
	\begin{minipage}{0.6\linewidth}\small
    \centering
		\includegraphics[width=\linewidth]{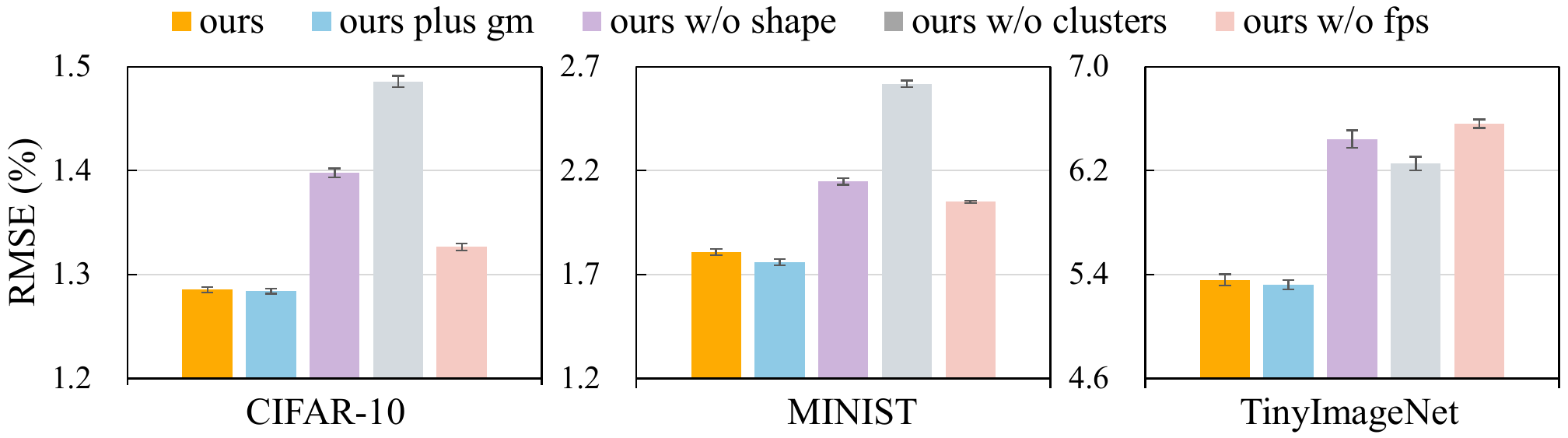}
		\captionof{figure}{Ablation and variant study on validation datasets. RMSE (\%) is reported: a shorter bar means a better representation. 
		``plus gm'': the global mean is additionally used. ``ours'': all the three representations are used for model accuracy prediction. 
		We observe none of the three representations is dispensable.} 
		\label{fig:fig-ablation}
	\end{minipage}\hfill
\end{table*}

\subsection{Main Evaluation}
\textbf{Comparison with two naive accuracy evaluators.} We first compare our method with two native methods: ``Prediction'' and ``Entropy'' in Table \ref{table:Result}. The former checks the maximum softmax score and uses a threshold to decide whether a sample is correctly classified. The latter considers the value calculated on the entropy of softmax outputs (normalized by $log(K)$), and data with a value less than a threshold is regarded as correct prediction. 
Results indicate that both methods are very sensitive to the threshold and are generally inferior to our method. For example, setting the threshold to 0.9 and 0.8 will lead to very different results. While getting sometimes good prediction performance on one dataset (\emph{e.g.,} CIFAR-C), the same threshold gives poor accuracy on other datasets (\emph{e.g.,} CIFAR-S). In fact, tuning the threshold as a hyperparameter is non-trivial, as different test sets would require different thresholds. Our observation is consistent with existing works \cite{deng2021labels,deng2021does}, and it is still an open question how to make them robust to thresholding. 

\textbf{Comparison with state-of-the-art methods.} We then compare our method with existing works in this field. They include average confidence (AC) and its associated measures \cite{guillory2021predicting}, the rotation prediction accuracy on the test set (``Rotation") \cite{deng2021does}, and the Fr\'{e}chet distance (FD) between training and test sets based on their difference in mean and variance \cite{deng2021labels}. We use published codes or try our best to implement their methods on these datasets. As Table \ref{table:Result} shows, we report \textbf{RMSE = 7.02\%, 9.52\% and 5.95\% on CIFAR-F, Digital-S and TinyImageNet-C}, respectively. These numbers consistently outperform the competing methods. For example, on Digital-S, our estimates are more accurate than AC \cite{guillory2021predicting} evidenced by a 1.83\% lower RMSE .

\textbf{Synthetic test sets \emph{vs.} real-world test sets.} When evaluating classifier accuracy, we use both synthetic test sets (\emph{e.g.,} CIFAR-C, and SVHN-C) and real test sets (\emph{e.g.,} CIFAR-F and Digital-S). In Table \ref{table:Result}, we observe that AutoEval on synthetic datasets is consistently better than real-world datasets. For example, RMSE on CIFAR-C can be as low as 1.28\%, while that on CIFAR-F is always higher than 7\%. Similar observation is also made on SVHN-C and Digital-S. That accuracy on human-made datasets is much easier to predict is probably because the sample sets used to train the regressor are also synthetic. Although we have made sure the sample sets have no overlapping image transformations with the test sets, they still possess some resemblance. In addition, for TinyImageNet-C, the lowest RMSE is 5.94\%, higher than those on the smaller CIFAR-C and SVHN-C. This is because the TinyImageNet setting is a 200-way classification problem, which is more challenging for the classifier. Meanwhile, TinyImageNet has more categories, so its distribution is more complex to be represented. 

\textbf{AutoEval for VGG-16.} We experimented not only with ResNet-44 (Table \ref{table:Result}), but also with VGG-16, and the results under the CIFAR-10 setup are shown in Table~\ref{table:vgg-16}. Compared with competing methods \cite{deng2021does,deng2021labels,guillory2021predicting}, our method is again superior.

\begin{figure*}[t]
\begin{center}
	\includegraphics[width=\linewidth]{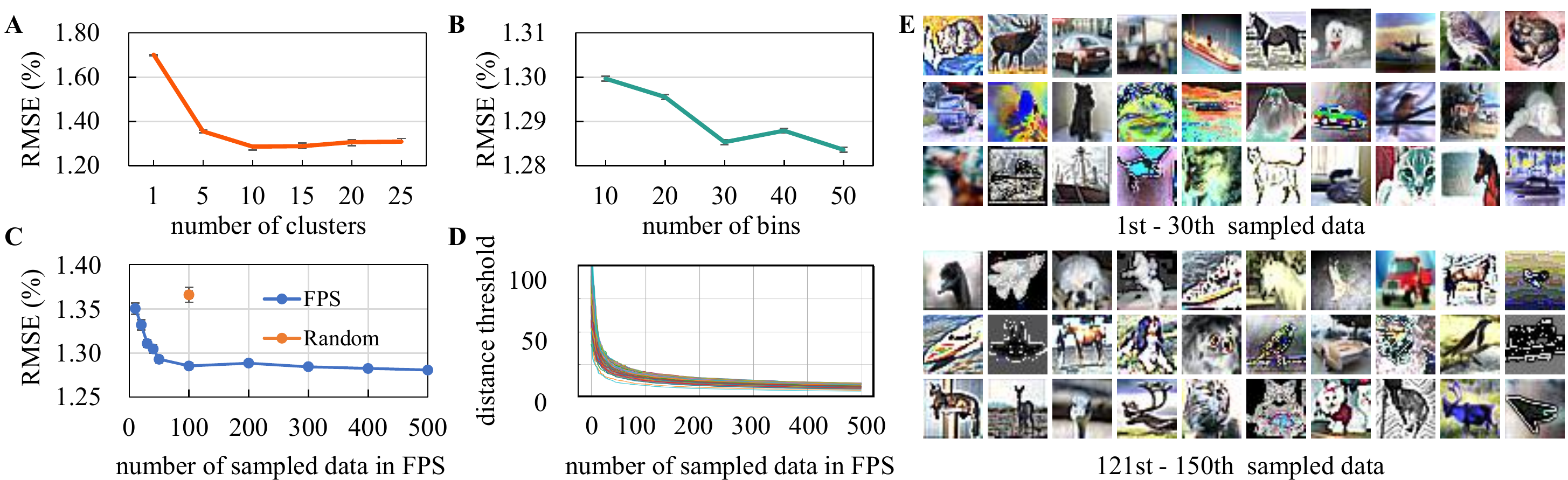}
\end{center}
\caption{Parameter analysis under the CIFAR-10 setup. \textbf{A}: number of clusters. \textbf{B}: number of bins in the feature histogram. \textbf{C}: number of images sampled by FPS. \textbf{D}: Relationship between number of sampled data and the distance threshold in FPS.  
Each curve in \textbf{D} represents a validation set under the CIFAR-10 setup. As number of the sample data increases, \textbf{C} and \textbf{D} exhibit a similar ``elbow'' shape. \textbf{E}: the first 30 images and images between 121st and 150th sampled by FPS. The first 30 images are much more diverse than images sampled later.} 
\label{fig:fig-ex-parameters}
\end{figure*}

\textbf{Ablation study: effectiveness of each of the three set features.} Fig.~\ref{fig:fig-ablation} summarizes the ablation study on the validation sets of each setup. We have the following two observations. 
First and most importantly, using the three features together (``ours'' in Fig.~\ref{fig:fig-ablation}) allows us to obtain more accurate estimates than  deleting any component. For example, on the validation set of the CIFAR-10 setup, removing the shape, clusters or FPS alone leads to an increase in RMSE of 0.0411\%, 0.2004\% and 0.1124\%, respectively. Our results clearly indicate that the three features underlying shape, internal distribution and boundary spread are complementary to each other and that none of them is dispensable in our system. Second, as a minor point, we observe that the effect of the three features might vary from dataset to dataset. For example, the cluster centers seem more important on CIFAR-10 than TinyImageNet. The reason is still not: it may be due to many reasons such as dataset complexity (\emph{e.g.,} number of classes) and image resolution, and it would be interesting to do further studies. 

\textbf{Impact of incorporating the global mean.} In Fig.~\ref{fig:fig-ablation}, when the global mean vector is added to our semi-structured representation, we observe very slight improvement on MNIST and TinyImageNet setups, but not on CIFAR-10. We speculate that the global mean cues can be well reflected by the local means, so its effect would be less prominent. 

\textbf{Comparison between FPS sampling and random sampling.} In FPS, the sampled points should be possibly far from each other, while in random sampling, the resulting points approximately follow the dataset distribution. In Fig.~\ref{fig:fig-ex-parameters} \textbf{C}, we compare FPS sampling with random sampling under 100 sampled data. Note that the RMSE of random sampling is averaged over 5 runs. We clearly observe that FPS sampling yields lower RMSE than random sampling, confirming the benefit of covering the spread of the distribution when performing sampling. 

\subsection{Parameter analysis}
The proposed semi-structured dataset representations have a few key hyperparameters, including the number of clusters, dimension of the feature histogram and number of images sampled by FPS. For all three, we use the validation sample sets
to select the best values. When tuning one hyperparameter, the value of the others are fixed at their near-optimal values. Below we show the hyperparameter tuning process under CIFAR-10 setup as an example.

\textbf{Number of clusters.} Our validation results under the CIFAR-10 setup are shown in Fig.~\ref{fig:fig-ex-parameters} \textbf{A}. The RMSE of the validation set is lowest when the number of clusters is $10$ or slightly higher. In fact, we find that setting the cluster number as the same value of the number of classes is desirable for all of the CIFAR-10, MNIST and TinyImageNet setups. This is reasonable because under the classification scenario, image features tend to exhibit clustering patterns. 

\textbf{Dimension of the histogram.} In Fig.~\ref{fig:fig-ex-parameters} \textbf{B}, we show the validation results of this hyperparameter, where we find the best value is about $30$. Similarly, the best values under the MNIST and TinyImageNet setups are 30 and 50, respectively.

\textbf{Number of images sampled by FPS.} As Fig.~\ref{fig:fig-ex-parameters} \textbf{C} presents, RMSE drops and then becomes relatively stable when the number of sampled images increases. We choose to sample 100 images per dataset under the CIFAR-10 setup. This value becomes 100 and 400 under the MNIST and TinyImageNet setups, respectively. Fig.~\ref{fig:fig-ex-parameters} \textbf{D} shows the relationship between the number of sampled data and the FPS distance threshold. We can find that this curve is in a similar shape as Fig.~\ref{fig:fig-ex-parameters} \textbf{C}, and the ``elbow'' point which is around 100 seems a good choice of this hyperparameter. 



\section{Discussion}
\label{sec:discussion}

\textbf{Fusion of existing dataset-level statistics.} It is interesting to see how existing works in AutoEval complement our system. To this end, we perform experiments to fuse the average confidence \cite{guillory2021predicting} and rotation \cite{deng2021does}, and results are shown in Table~\ref{table:ours_fusion}. We observe that adding these statistics does not yield noticeable system improvement. It is probably because both statistics are 1-dim scalars, and, without a huge number of training sample sets, their semantics are obscured by our representations that are of a much higher dimension. That said, it would be of interest to study
fusion methods to combine existing findings into more discriminative representations, and we will study it in further work.

\textbf{Dataset-level representation and similarity.} This work conducts analysis on the dataset level. In analogy to image-level analysis, we could have not only dataset representations but also dataset-dataset similarities. There exist some hand-crafted similarity measurements on the dataset level, such as MMD and FD. This paper explores a more task-specific alternative, which is to learn and represent dataset similarities through AutoEval: the test set is similar to training set if the estimated test accuracy is high, and vice versa.


\begin{table}[t]
\begin{center}
\small
\setlength{\tabcolsep}{2mm}{
		\begin{tabular}{l | c| c| c}
        			\toprule
        Method		  & CIFAR-10.1 & CIFAR-C &  CIFAR-F\\
        			\midrule
Ours	& 0.6900 & 1.1830 & 6.5472 \\
Ours + AC~\cite{guillory2021predicting}	& 0.7100 & 1.1827 & 6.5140 \\
Ours + Rotation\cite{deng2021does}	& 0.7200 & 1.1832& 6.5503 \\
			\bottomrule
		\end{tabular}}
	\end{center}
	    \caption{AutoEval performance (RMSE, \%) of fused dataset-level statistics. The VGG-16 classification model is used. No obvious improvement is observed after fusion.}	\label{table:ours_fusion}
	    \vspace{-3mm}
\end{table}

\section{Conclusion}
We investigate the label-free model evaluation problem in this paper. It is motivated by inadequate discriminative power of existing approaches.  We propose a novel semi-structured representation for a new test set, which utilizes {\em feature histograms}, {\em locality-aware means}, as well as {representative samples with FPS}.  To enable AutoEval, the representations of sample set and the training set are fed into a regression network supervised by the classifier accuracy of the sample set.  Our method has achieved a consistent performance boost on a series of benchmarks and setups.  This work will hopefully shed light on the important but previously underexplored dataset-level analysis.

{\small
\bibliographystyle{ieee_fullname}
\bibliography{319}
}

\end{document}